\documentclass[letterpaper, 10 pt, conference]{ieeeconf}  

\IEEEoverridecommandlockouts                              

\usepackage{amsmath} 
\usepackage{amssymb}  

\usepackage{graphicx}
\usepackage[tight,footnotesize]{subfigure}
\usepackage{amsmath} 
\usepackage{amssymb}  
\usepackage{mathtools}
\usepackage{hyphenat}
\usepackage[table]{xcolor}
\usepackage{caption}
\usepackage{multirow}
\usepackage{url}
\usepackage{booktabs,amsfonts,dcolumn}
\usepackage{tabularx}
\usepackage{gensymb}
\usepackage{makecell}
\usepackage{cite}

\urldef{\mailmd}\path| md.modasshir@outlook.com|
\urldef{\mailsb}\path| yiannisr@cse.sc.edu|

\newcommand\bcomment[1]{\textcolor{red}{bjoshi:#1}}
\newcommand\scomment[1]{\textcolor{blue}{SR:#1}}
\newcommand\ycomment[1]{\textcolor{magenta}{IR:#1}}
\newcommand\hcomment[1]{\textcolor{green}{H:#1}}

\usepackage{xspace}

\long\def\invis#1{}

\newcommand\eq[1]{Eq.~\eqref{#1}}
\newcommand\fig[1]{Fig.~\ref{#1}}

\newcommand\tab[1]{Table~\ref{#1}}
\newcommand\alg[1]{Algorithm~\ref{#1}}

\makeatletter
\DeclareRobustCommand\onedot{\futurelet\@let@token\@onedot}
\def\@onedot{\ifx\@let@token.\else.\null\fi\xspace}

\def\eg{\emph{e.g}\onedot}

\def\etal{\emph{et al}\onedot}
\makeatother

\def\inv{^{-1}}

\usepackage[group-separator={,}]{siunitx}

\title{\LARGE \bf SM/VIO: Robust Underwater State Estimation\\
Switching Between Model-based and Visual Inertial Odometry\vspace{-0.1in}  }
\author{Bharat Joshi$^{a*}$, Hunter Damron$^{b*}$, Sharmin Rahman$^a$, and Ioannis Rekleitis$^a$\vspace{-0.1in}%
\thanks{$^*$ The first two authors have contributed equally to the paper.}%
\thanks{$^a$University of South Carolina, Columbia, SC, USA, yiannisr@cse.sc.edu, \{bjoshi,srahman\}@email.sc.edu.} %
\thanks{$^b$Epic Systems Corporation, Madison, WI, USA, 53703, {\tt\small hdamron@epic.com}}%
\thanks{The authors would like to acknowledge the generous support of the National Science Foundation grants (NSF 2024741, 1943205).}%
}
\begin{document}

\begin{minipage}{0.90\textwidth}\ \\[12pt]  
\begin{center}
     This paper has been accepted for publication in \textit{IEEE Conference on Robotics and Automation 2023}.  
\end{center}
  \vspace{1in}
  ©2023 IEEE. Personal use of this material is permitted. Permission from IEEE must be obtained for all other uses, in any current or future media, including reprinting/republishing this material for advertising or promotional purposes, creating new collective works, for resale or redistribution to servers or lists, or reuse of any copyrighted component of this work in other works.
\end{minipage}

\newpage

\maketitle              
\thispagestyle{empty}
\pagestyle{empty}

\begin{abstract}
    This paper addresses the robustness problem of visual-inertial state estimation for underwater operations. Underwater robots operating in a challenging environment are required to know their pose at all times\invis{, otherwise they will be lost}. All vision-based localization schemes are prone to failure due to poor visibility conditions, color loss, and lack of features. The proposed approach utilizes a model of the robot's kinematics together with proprioceptive sensors to maintain the pose estimate during visual-inertial odometry (VIO) failures. Furthermore, the trajectories from successful VIO and the ones from the model\hyp driven odometry are integrated in a coherent set that maintains a consistent pose at all times. Health\hyp monitoring tracks the VIO process ensuring timely switches between the two estimators. Finally, loop closure is implemented on the overall trajectory. The resulting framework is a robust estimator switching between model\hyp based and visual-inertial odometry (SM/VIO). Experimental results from numerous deployments of the Aqua2 vehicle demonstrate the robustness of our approach over coral reefs and a shipwreck.   
\end{abstract}

\section{INTRODUCTION}

\invis{
Seventy one percent of the Earth is covered by water~\cite{USGS71} and the role of the aquatic environment is crucial in several aspects. The oceans play a critical role in climate control; the coral reefs, while fragile and vulnerable to rising temperatures~\cite{Hodgson2002ReefCrisis,Mulhall2007Reef}, contain the highest biodiversity on the planet~\cite{1994ReefMonitorManual}; coastal zones are often hosts of numerous economic activities; while ports are among the most critical infrastructures for trade and transportation. Historical shipwrecks tell an important part of history and at the same time have a special allure for most humans, as exemplified by the plethora of movies and artworks of the Titanic; see, e.g.,~\cite{eustice2006visually} for the visual mapping of the Titanic. Shipwrecks are also one of the top scuba diving attractions all over the world.
}

\begin{figure}[t]
 \centering
{\includegraphics[width=0.85\columnwidth]{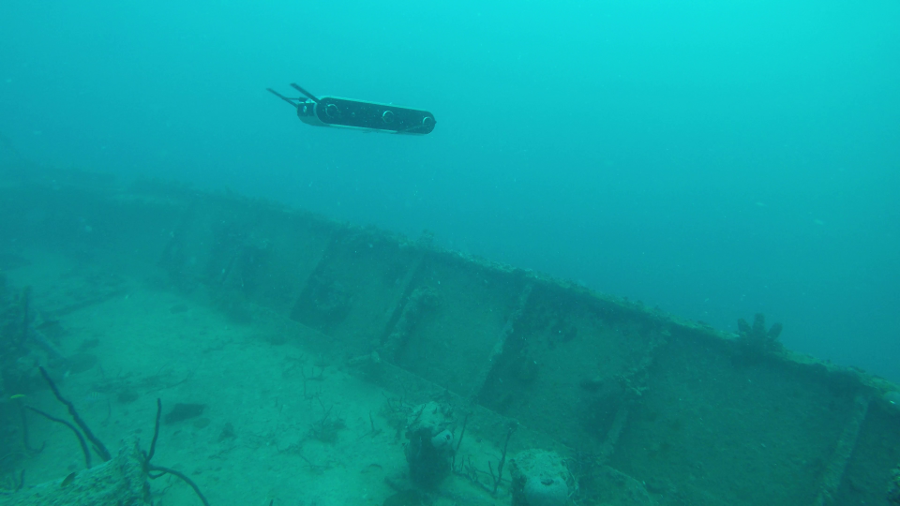}}
\caption{Aqua2 AUV navigating over the Stavronikita shipwreck, Barbados. The front cameras are only seeing blue water when approaching the side of the wreck.}\vspace{-0.1in}
\label{fig:beauty}
\end{figure}

This paper proposes a novel framework for solving the robustness problem of state estimation underwater. Central to any autonomous operation is the ability of the robot to know where it is with respect to the environment, a task described under the general term of state estimation. Over the years many different approaches have been proposed; however, state estimation underwater is a challenging problem that still remains open. Vision provides rich semantic information and through place recognition results in loop closures. Unfortunately, as demonstrated in recent work on comparing numerous open\hyp source packages of visual and visual/inertial state estimation~\cite{QuattriniLiISERVO2016,JoshiIROS2019}, in an underwater environment there are frequent failures for a variety of reasons. In contrast to above water scenarios, GPS-based localization is impossible. In addition to the traditional difficulties of vision-based localization, the underwater environment is prone to rapid changes in lighting conditions, limited visibility, and loss of contrast and color information with depth. \invis{More specifically, the presence of caustic patterns, which are due to refraction, non-uniform reflection, and penetration of light when it transitions from air to the rippling surface of the water changes the light intensity.} Light scattering from suspended plankton and other matter causes ``snow effects''  and blurring, while the incident angle at which light rays hit the surface of the water can change the visibility at different times of the day~\cite{2007light}. Finally, as light travels at increasing depths, different parts of its spectrum are absorbed; red is the first color that is seen as black, and eventually orange, yellow, green, and blue follow~\cite{SkaffBMVC2008,roznere2019color}. In addition to all the above underwater specific challenges, an unknown environment often presents areas where there are no visible landmarks. For example, in \fig{fig:beauty} an Aqua2~\cite{DudekIROS2005} Autonomous Underwater Vehicle (AUV) mapping the deck of a shipwreck reaches the starboard side where the front cameras see only empty water with no features. 

Visual inertial odometry (VIO) has been used for state estimation in a multitude of environments such as indoor, outdoor and even gained some traction in harsh environments such as underwater~\cite{RahmanIJRR2022}. While most VIO research often focuses on improving accuracy, robustness is as critical for autonomous operations. If VIO fails during deployment the results could be catastrophic leading to vehicle loss. From our early investigations~\cite{QuattriniLiISERVO2016,JoshiIROS2019}, many vision-based approaches diverge, or outright fail, sometimes at random;  however, deploying a vehicle  underwater in autonomous mode requires that it will return to base, or a collection point, during every deployment. It is very important for AUVs to be able to keep track of their pose; even with diminished accuracy; over the whole  operation. We propose switching between VIO and a model-based  estimator addressing the accuracy and robustness of state estimation by identifying failure modes, generating robust predictors for estimator divergence/failure, always producing a pose estimate.   

\invis{While state estimation research often focuses on improving accuracy, robustness is as critical for autonomous operations. If an estimator fails during deployment the results could be catastrophic leading to vehicle loss. From our early investigations~\cite{QuattriniLiISERVO2016,JoshiIROS2019}, many approaches diverge, or outright fail, at random when run under exactly the same conditions on the same dataset;  however, deploying a vehicle  underwater in autonomous mode requires that it will return to base, or a collection point, during every deployment.  Furthermore, AUVs have limited computing power and any estimator has to take that into consideration. The proposed approach addresses the accuracy and robustness of state estimation by identifying failure modes, generating robust predictors for estimator divergence/failure, producing a pose estimate at all times\invis{, and making contributions to the theory of switching observers (estimators) in a robust manner.}}  

The core of the proposed approach is a robust switching estimator framework, which always provides a realistic estimate reflecting the true state of the vehicle. First of all, the health of  VIO~\cite{RahmanICRA2018,RahmanIROS2019a} is monitored by tracking the number of features detected, their spatial distribution, their quality, and their temporal continuity. By utilizing the measures described above when an estimator starts diverging, before complete failure, an alternative estimator is introduced based on sensor inputs robust to underwater environment changes. For example, there is a model-based estimator~\cite{6942867,meger2015learning} used for controlling the Aqua2 vehicles combining the inertial and water depth signals together with the flipper configuration and velocity~\cite{plamondon2008trajectory,giguere2006characterization}; when the visual/inertial input deteriorates, the proposed system switches to the model-based estimator until the visual/inertial estimates are valid again. The choice of switching-based loosely coupled fusion of odometry estimates ensures flexibility in choosing both the VIO and the conservative estimator in a modular fashion. The two estimators switch back and forth based on the health status of the VIO estimator. Finally, a loop\hyp closure framework ensures the consistent improvement of the combined estimator. Our main contribution is a robust switching-based state estimation framework termed Robust Switching Model-based/Visual Inertial Odometry (SM/VIO) capable of keeping track of an AUV even when VIO fails. This allows the AUV to carry out underlying tasks such as path planning, coverage, and performing motion patterns maintaining a steady pose and relocalize when visiting previous areas. Extensive experiments over different terrains validate the contribution of the proposed robust switching estimator framework in maintaining a realistic pose of the AUV at all times. In contrast, state-of-the-art VIO algorithms~\cite{RahmanIJRR2022, leutenegger2015keyframe, qin2019b, Geneva2020ICRA, campos2021orb} result in a much higher error or even complete failure.

\section{RELATED WORK}\label{sec:related}
\invis{-0.1in}
In recent years a plethora of open source packages addressing the problem of vision-based state estimation has appeared~\cite{ShkurtiIROS2011,klein,Newcombe,raey,forster2014svo,ball2013openratslam,Davison2007,7219438,Mur-Artal-RSS-15,Geneva2020ICRA,RahmanICRA2018,RahmanIROS2019a}.  Quattrini Li \etal \cite{QuattriniLiISERVO2016} compared several packages on a variety of datasets  to measure the performance in different environments.  Extending the above comparison with a focus on the underwater domain, Joshi \etal~\cite{JoshiIROS2019} investigated the performance of VIO  packages. The above comparisons demonstrated that many packages require special motions~\cite{klein}, or only work for a limited number of images~\cite{lour09,zhao2015parallaxba}, or are strictly offline~\cite{schoenberger2016sfm}. Furthermore, intermittent failures were observed, the most common explanation being the random nature of the RANSAC technique~\cite{fischler1981random} utilized by most of them. The underwater state estimation approach SVIn2 by Rahman \etal~\cite{RahmanIJRR2022}  demonstrated improved accuracy and robustness; however, it did not provide any assurances for uninterrupted estimates, which is the focus of this paper.

Utilizing an AUV to explore an underwater environment has gained popularity over the years. Sonar and stereo estimation for object modeling has been proposed in~\cite{6741221,6224731}. Nornes \etal~\cite{nornes2015} acquired stereo images utilizing an ROV off the coast of Trondheim Harbour, Norway. \invis{A commercially available software has been used to process the images to reconstruct a model of the shipwreck.} In~\cite{soreide2005} a deep\hyp water ROV is adopted to map, survey, sample, and excavate a shipwreck area. Sedlazeck \etal~\cite{sedlazeck2009}, reconstructed a shipwreck in 3D by pre\hyp processing images collected by an ROV and applying a Structure from Motion based algorithm. The  images  used for testing such an algorithm contained some structure and  a  lot  of background,  where  only  water  was  visible. Submerged structures were reconstructed in  3D~\cite{5649213}. Finally, recent work by Nisar \etal~\cite{8721075} proposed the use of a model-based estimator to calculate external forces in addition to the pose of aerial vehicles, ignoring failure modes of VIO. In all previous work, when the state estimation failed there was no recovery. In contrast, the proposed approach of SM/VIO for underwater environments addresses the VIO failure and the AUV can continue operations until reaching another feature\hyp rich area.

The use of switching estimators (also called observers) has not been applied in many mobile robotics applications and not, to our knowledge, to an AUV. Liu~\cite{650020} presented a generic approach for non-linear systems. Suzuki \etal~\cite{8722845} utilized a switching observer to model ground properties together with the robot's kinematics. Manderson \etal~\cite{Manderson2020rss} utilized a model estimator in conjunction with Direct Sparse Odometry~\cite{engel2017direct}  without monitoring the health, switching estimators, and merging the two trajectories into one.

\section{THE PROPOSED SYSTEM}\label{sec:proposed}
\begin{figure}[t]
 \centering
 \vspace{0.1in}
\fbox{\includegraphics[width=0.9\columnwidth]{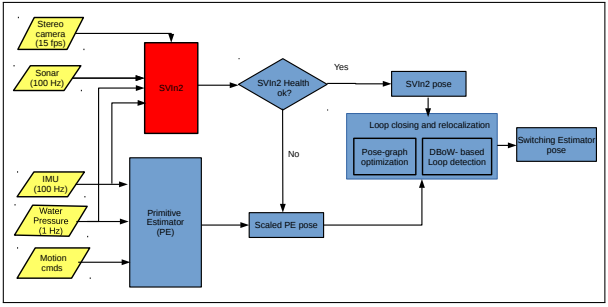}}
\caption{Overview of the switching estimator.\invis{-0.2in}}
\label{fig:diagram}
\end{figure}

\begin{figure*}
    \centering
    \begin{tabular}{ccc}        
        \subfigure[]{\includegraphics[height=0.2\textheight]{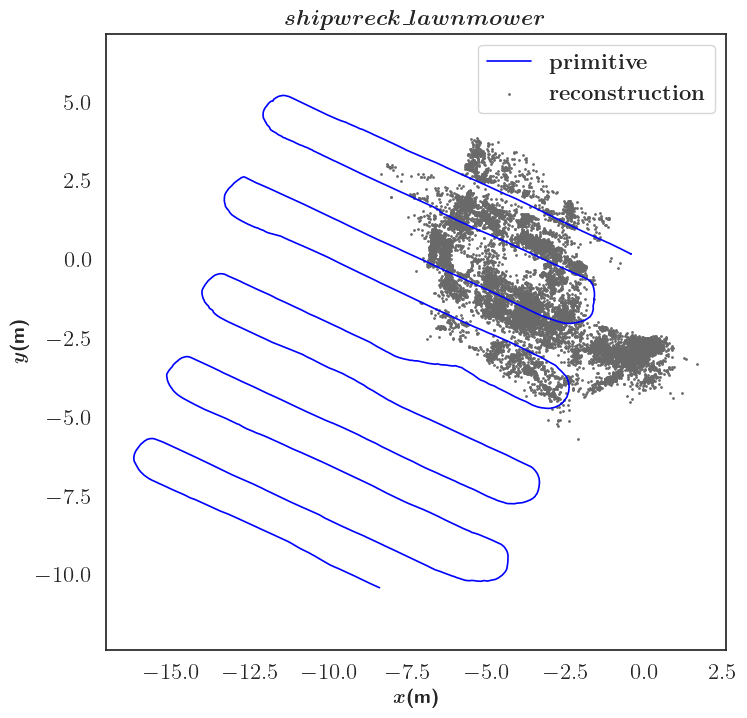}}&
        \subfigure[]{\includegraphics[height=0.2\textheight]{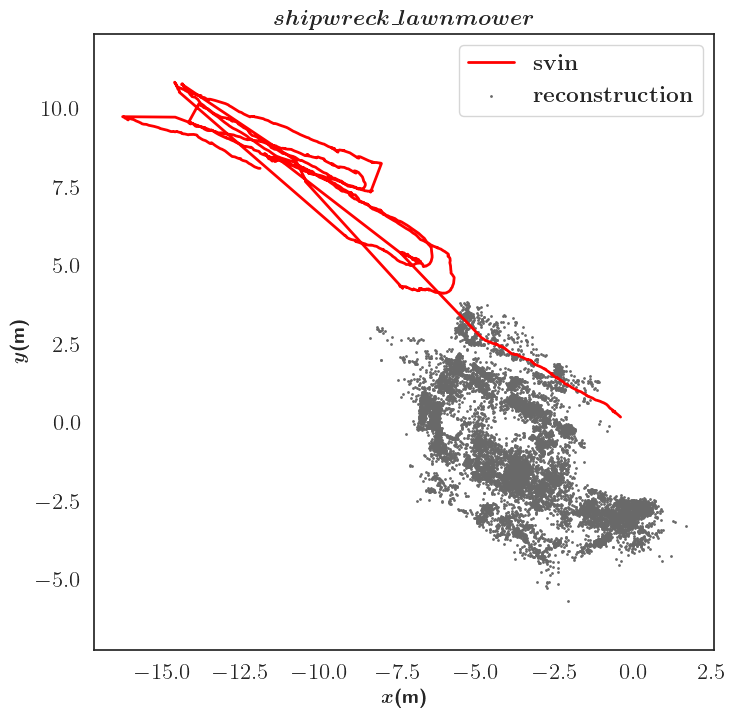}}&
        \subfigure[]{\includegraphics[height=0.2\textheight]{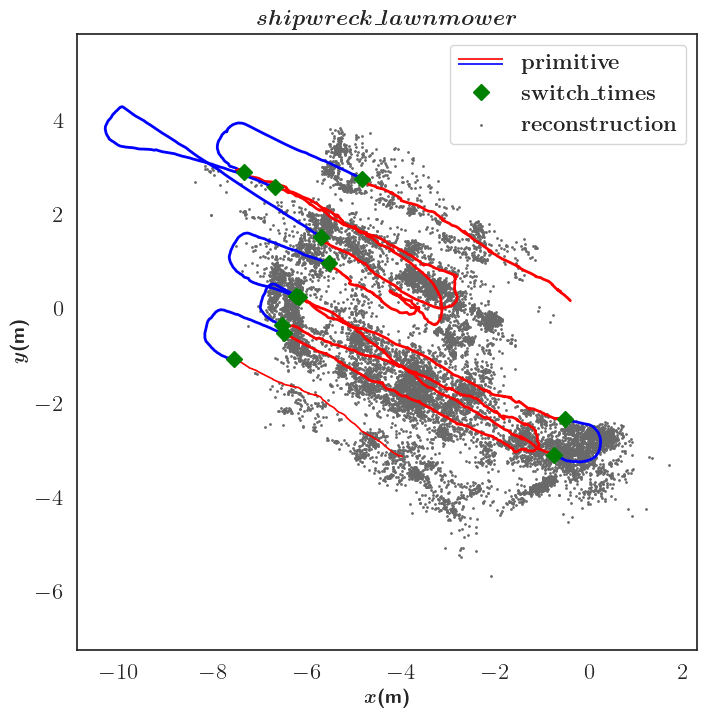}} \\
        \vspace{-0.1in}\subfigure[]{\includegraphics[height=0.15\textheight]{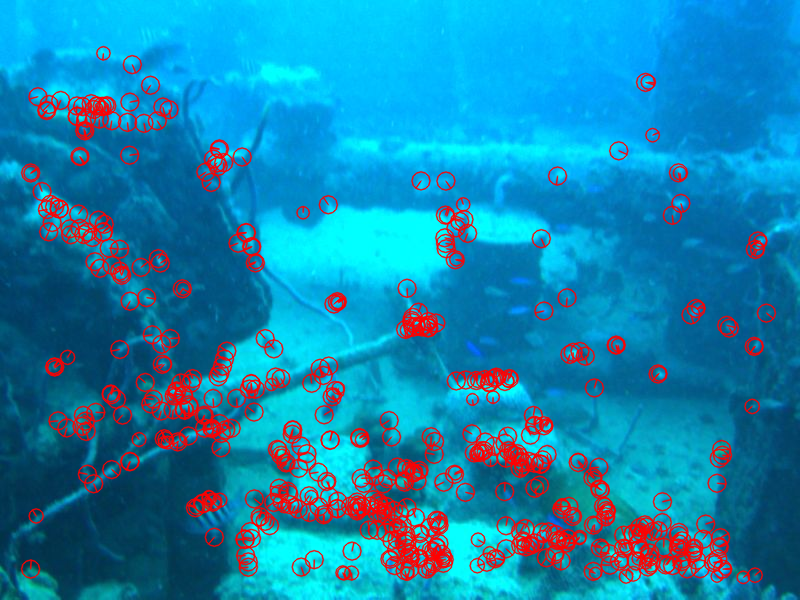}}&
        \subfigure[]{\includegraphics[height=0.15\textheight]{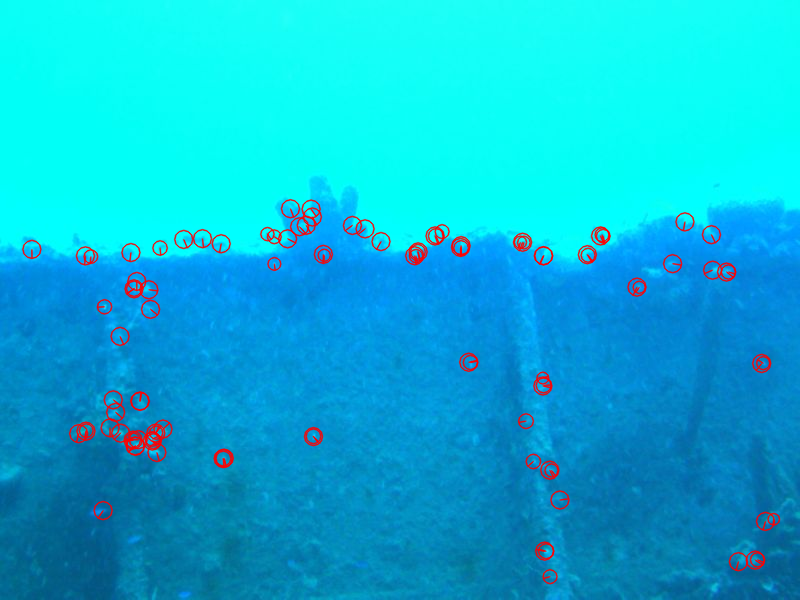}}&
        \subfigure[]{\includegraphics[height=0.15\textheight]{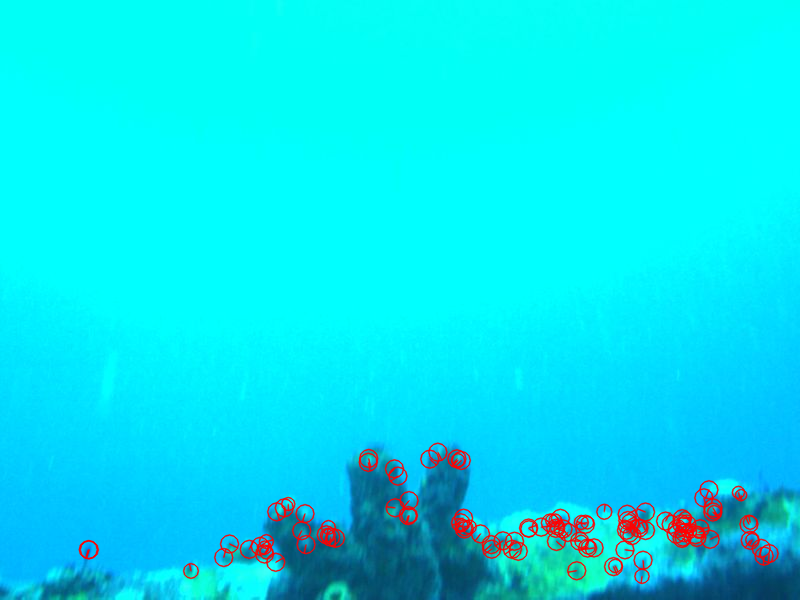}}    \end{tabular}
    \vspace{-0.0in}\caption{First row, an overview example: (a) Trajectory according to the primitive estimator; \invis{using the same algorithm as the robot controller} PE believes the AUV performed a near perfect lawnmower pattern. (b) Trajectory according to SVIn2~\cite{RahmanIROS2019a}; due to tracking loss the VIO is way off the actual wreck. (c) Trajectory resulting from the proposed method; the switching estimator utilized the robust parts of VIO (in red) switching to PE when tracking was lost (in blue). The stable 3D features detected are plotted as grey points. Second row, characteristic images of the shipwreck: (d) the AUV is over the wreck seeing the deck; (e) the AUV is approaching the side of the wreck, still able to localize, but the number of features decreases; (f) the AUV is at the edge of the wreck seeing mostly blue water and the estimator switches from VIO to PE.}
    \label{fig:example}
\end{figure*}

\paragraph{\bf Overview} The proposed approach (SM/VIO) utilizes a model\hyp based estimator termed primitive estimator (PE), utilizing the water depth sensor, the IMU, and the motor commands to propagate the state of the AUV forward when the visual-inertial estimator fails; see \fig{fig:example}(a) for an estimate from PE. It is worth noting that the AUV is using the same model to navigate, as such the PE estimate of the lawnmower pattern in \fig{fig:example}(a) follows the exact pattern, however, does not correspond to the actual trajectory which was affected by external forces (e.g. water current). When VIO is consistent it is the preferred estimator having higher accuracy due to the exteroceptive sensors (vision and acoustic). Key to the proposed approach is a health monitor process that tracks the performance of VIO over time and informs a decision for switching between  VIO and PE; see \fig{fig:example}(c) for the switching estimator trajectory, where the switch points are marked green. When the VIO restarts tracking successfully, the health monitor informs the switch from the PE to the VIO. Throughout this process a consistent pose is maintained. More specifically, when the VIO fails, the PE is initialized with the last accurate pose from VIO, and when the VIO restarts the last pose of PE is utilized. Finally, during VIO controlled operations, loop closure is performed, also optimizing  the PE produced trajectories; the complete framework is outlined in \fig{fig:diagram}. Following the approach of Joshi \etal~\cite{JoshiICRA2022}, the stable 3D features are tracked and their position is updated after every loop closure, thus resulting into a consistent point cloud. Next we discuss the individual components of SM/VIO.

\invis{More complex is the situation when VIO restarts, in addition to initializing to the respective pose produced by the PE, the trajectory of poses produced by the PE is added to the graph optimization framework of VIO.}

The target vehicle is the Aqua2 AUV~\cite{DudekIROS2005}, an amphibious hexapod robot.\invis{, approximately $65{\rm cm} \times 45{\rm cm} \times 13{\rm cm}$ in size and $10{\rm kg}$ in weight.} Underwater, Aqua2 utilizes the motion from six flippers, each actuated independently by an electric motor.\invis{, to move in 3D. Aqua2 has 6DoF, of which five are controllable, two directions of translation (forward/backward and upward/downward), along with roll, pitch and yaw. 
}
The robot's pose is described using the vector $\textbf{x} = \begin{bmatrix}
     _{W}\textbf{p}_{I}^{T},&
     _{W}\textbf{q}_{I}^{T}
\end{bmatrix},$
\invis{\begin{equation}
\label{eq:state}
\textbf{x} = \begin{bmatrix}
     _{W}\textbf{p}_{I}^{T},&
     _{W}\textbf{q}_{I}^{T}
\end{bmatrix},
\end{equation}}
 $_{W}\textbf{p}_{I}^{T}=[x, y, z]$ represents the position of the robot in the world frame, and $_{W}\textbf{q}_{I}^{T}=[q_w, q_x, q_y, q_z]$ is the quaternion representing the robot's attitude. \invis{The robot's computational system consists of two units, one responsible for vision and high-level planning and the other responsible for control related computations.}
Aqua2 vehicles are equipped with three cameras, an IMU, and a water pressure sensor.

\invis{Aqua2 vehicles are equipped with three \invis{iDS USB 3.0 UEye }cameras, two facing forward and  one in the back. \invis{The forward-facing cameras are used for navigation and data collection.}
The fields of view of the cameras are $120^\circ$ (horizontal) and $90^\circ$ (vertical) tilted downward by $40^\circ$ from the horizontal plane. Additional sensors include an IMU and a water pressure sensor which are used for motion control. }

\paragraph{\bf Primitive Estimator}
The primitive estimator maintains a local copy of the robot's pose $[{}_W\textbf p_I^T,{}_W\textbf q_I^T]$, which is updated at a rate of \SI{100}{\Hz}.
The IMU provides an absolute measurement of $_W\textbf q_I^T$.
The velocity of the robot is estimated by the forward speed command $v_{x}$ and the heave (up/down) command $v_{z}$ sent to the Aqua2 during field trails. These commands are used by Aqua2 to perform motion primitives and control the flipper motion. Since the same commands are used for Aqua2 control and PE predictions, the resulting PE trajectories will look perfectly aligned with the desired motion primitives; this is a drawback of just using PE prediction. 
At each time step $t$, the position is updated by
\begin{equation}
\label{eq:prim_estimator_update}
    {}_W\textbf p_{I,t+1} \coloneqq {}_W\textbf p_{I,t} + {}_W\textbf{R}_{I,t} [v_x^t, 0, v_z^t]^T \Delta{t}
\end{equation}
where $_W\textbf{R}_{I}$ is the rotation matrix corresponding to $_W\textbf{q}_{I}$.
Because the water pressure sensor provides an absolute measurement of depth, this measurement is used instead of the above estimate for $z$. Moreover, the forward velocity estimates are correct only up to scale depending on external forces (\eg ocean currents) and acceleration measurements error accumulation. Hence, before integrating the PE trajectory into the robust switching estimator, we scale the PE trajectory using the scaling factor between the PE and the VIO trajectory, as explained later.

\paragraph{\bf SVIn2 Review}
We use a VIO system that fuses information from  visual, inertial, water pressure (depth), and acoustic (sonar) sensors presented in Rahman \etal~\cite{RahmanIJRR2022,RahmanICRA2018,RahmanIROS2019a}, termed SVIn2. More specifically, SVIn2 estimates the state  of the robot by minimizing a joint estimate of the reprojection error and the IMU error, with the addition of the sonar error and the water depth error. SVIn2 performs non-linear optimization on sliding-window keyframes using the reprojection error and the IMU error term formulation similar to Leutenegger \etal~\cite{leutenegger2015keyframe}. \invis{The sonar range error \cite{RahmanICRA2018} represents the difference between the 3D point that can be derived from the range measurement and a corresponding visual feature in 3D.} The depth error term can be calculated as the difference between the AUV's position along the $z$ direction and the  water depth measurement provided by a pressure sensor. 

\emph{Loop-closing} and \emph{relocalization} is achieved using the binary bag-of-words place recognition module DBoW2 \cite{galvez2012bagofwords}. The loop closure module maintains a pose graph with odometry edges between successive keyframes and a loop-closure odometry edge is added between the current keyframe and a loop closure candidate when they have enough descriptor matches and pass PnP RANSAC-based geometric verification. For a complete description, please refer to~\cite{RahmanIJRR2022}.

\paragraph{\bf Health Monitoring}
As described in earlier studies~\cite{QuattriniLiISERVO2016,JoshiIROS2019}, estimators often diverge or outright fail even in conditions where they were working before;  intermittent failures are much more challenging in the field. Robustness measures and divergence predictors are crucial in detecting imminent failures. To monitor the health of the vision-based state estimator, we employ the following criteria hierarchically; the most important criterion is checked first. The VIO health is evaluated based on the following conditions hierarchically and considered untrustworthy based on:

\begin{enumerate}
    \item Keyframe detection. If a keyframe has not been detected after $kf\_wait\_time$ seconds the VIO has failed. The only exception is when the system is stationary (zero velocity). 
    \item The number of triangulated 3D keypoints that have feature detections in the current keyframe is less than a specified threshold, \textit{min\_kps}. We found that \textit{min\_kps} between 10-20 worked well. 
    \item The number of feature detections per quadrant, in the current keyframe, is less than a specified threshold, \textit{min\_kps\_per\_quadrant}. To account for situations where there are high number of features detected robustly in a small area; see \fig{fig:example}(e-f) where the bottom two quadrants contain all the features. The quadrant criterion is applied only if the total number of feature detections is less than $10 \times min\_kps\_per\_quadrant$.
    \item The ratio of new keypoints to the total keypoints is more than 0.75. The newly triangulated points are those that were not observable previously.
    \item The ratio of keypoints with feature detector response less than the average feature detector response in the current keyframe to the total keypoints is more than 0.85. The choice of a high threshold for the ratio is motivated by the fact that hierarchically more important criteria have already been satisfied. Hence, this criterion has less importance overall.
\end{enumerate}

Please note, the choice of the above parameters is flexible. For instance, the minimum number of tracked keypoints should be higher than the minimum number of points required for relative camera pose estimation using epipolar geometry. \invis{In addition, the number of new keypoint triangulations for successful tracking is a small number.} Thus, these parameters should only be taken as reference. During our experiments, we found out that changing the parameters slightly does not change the performance of the switching estimator greatly and the parameters where selected through experimental verification.

\paragraph{\bf Integration of SVIn2 and Primitive Estimator results} \label{subsec:integration}
Utilizing the framework described in Rahman \etal~\cite{RahmanICRA2018,RahmanIROS2019a} the graph SLAM formulation, based on the Ceres package~\cite{ceres}, is augmented to consider estimates from multiple observers thus maintaining the history of the estimates and enabling loop closures.\invis{ when revisiting the same place. }

\invis{It might be seen as SVIn2 cannot track again after divergence. May need to rewrite. It is worth noting that most state estimation packages, upon divergence, they cannot recover.}

We denote the poses SVIn2 and PE as $_W\textbf{T}_{sv}$ and $_W\textbf{T}_{pe}$, respectively, representing them as homogeneous $4\times4$ transformation matrices. The goal of the integration process is to provide a robust switching estimator pose $_W\textbf{T}_{ro}$ which matches $_W\textbf{T}_{sv}$ locally when SVIn2 is properly running, and matches $_W\textbf{T}_{pe}$ locally when SVIn2 is reporting failure. To find the scaling factor between SVIn2 and PE, we compute the ratio of the two trajectory lengths when both estimators are tracking well. More specifically, we compute the relative distance travelled as estimated by PE and SVIn2 between successive keyframes at time $t$ and $t+1$ and compute the scaling factor $s$ as:

\begin{equation}
    s = \frac{ \sum
    {\lVert {}_{W}\textbf{R}_{sv,t}\inv ({}_{W}\textbf{P}_{sv,t+1} - {}_{W}\textbf{P}_{sv,t}) \rVert}}
    {\sum {\lVert {}_{W}\textbf{R}_{pr,t}\inv ({}_{W}\textbf{P}_{pr,t+1} - {}_{W}\textbf{P}_{pr,t}) \rVert}}
\end{equation}

The scaling factor keeps updating over time whenever SVIn2 is tracking, to account for any changes in external factors. For the sake of convenience, we assume that the PE pose $_W\textbf{T}_{pe}$ is appropriately scaled by the scaling factor, $s$, to match the SVIn2 scale. Initially, when SVIn2 starts tracking, $_W\textbf{T}_{sv}$ is equivalent to  $_W\textbf{T}_{ro}$. When SVIn2 fails, we keep track of robust estimator pose $_W\textbf{T}_{ro}^{st}$ and primitive estimator pose  ${}_W\textbf{T}_{pe}^{st}$ at switching time, $st$. When PE is working normally, we compute the relative displacement of the current PE pose with respect to PE pose at the time of switching by ${}_W\textbf{T}_{pe}^{st\inv} \cdot {}_W\textbf{T}_{pe}$. This local displacement is then applied to the robust estimator pose using Eq. \ref{eq:tf_switch_prim} while making sure that the robust estimator pose tracks the PE pose locally during this time.

\begin{equation}
\label{eq:tf_switch_prim}
    {}_W\textbf{T}_{ro}  \coloneqq {}_{W}\textbf{T}_{ro}^{st} \cdot {}_W\textbf{T}_{pe}^{st\inv} \cdot {}_W\textbf{T}_{pe}
\end{equation}

It should be noted that ${}_{W}\textbf{T}_{ro}^{st} 
 \cdot {}_W\textbf{T}_{pe}^{st\inv}$ remains constant until SVIn2 starts tracking again. 

Similarly, when switching from the primitive estimator back to SVIn2, the robust estimator tracks the local displacement from SVIn2 using Eq. \ref{eq:switch_svin} with ${}_W\textbf{T}_{sv}^{st}$ remaining constant until next switch to PE occurs.

\begin{equation}
    \label{eq:switch_svin}
    {}_W\textbf{T}_{ro} \coloneqq {}_{W}\textbf{T}_{ro}^{st} \cdot {}_W\textbf{T}_{sv}^{st\inv} \cdot {}_W\textbf{T}_{sv}
\end{equation}

We make sure that the robust estimator tracks PE locally when SVIn2 fails and tracks SVIn2 again when it recovers as VIO is the preferred estimator maintaining robust uninterrupted pose estimate. As SVIn2 is capable of maintaining an accurate estimate in the presence of brief failures of visual tracking by relying on  inertial data, it is not desirable to switch between SVIn2 and PE back and forth frequently, as this introduces additional noise.  To reduce frequent switching between estimators, we wait for a small number of successive tracking failures to switch from SVIn2 to PE and vice-versa.
 \begin{figure*}[h!]
    \centering
    \vspace{0.05in}
    \begin{tabular}{ccc}        
        \subfigure[]{\includegraphics[height=0.13\textheight]{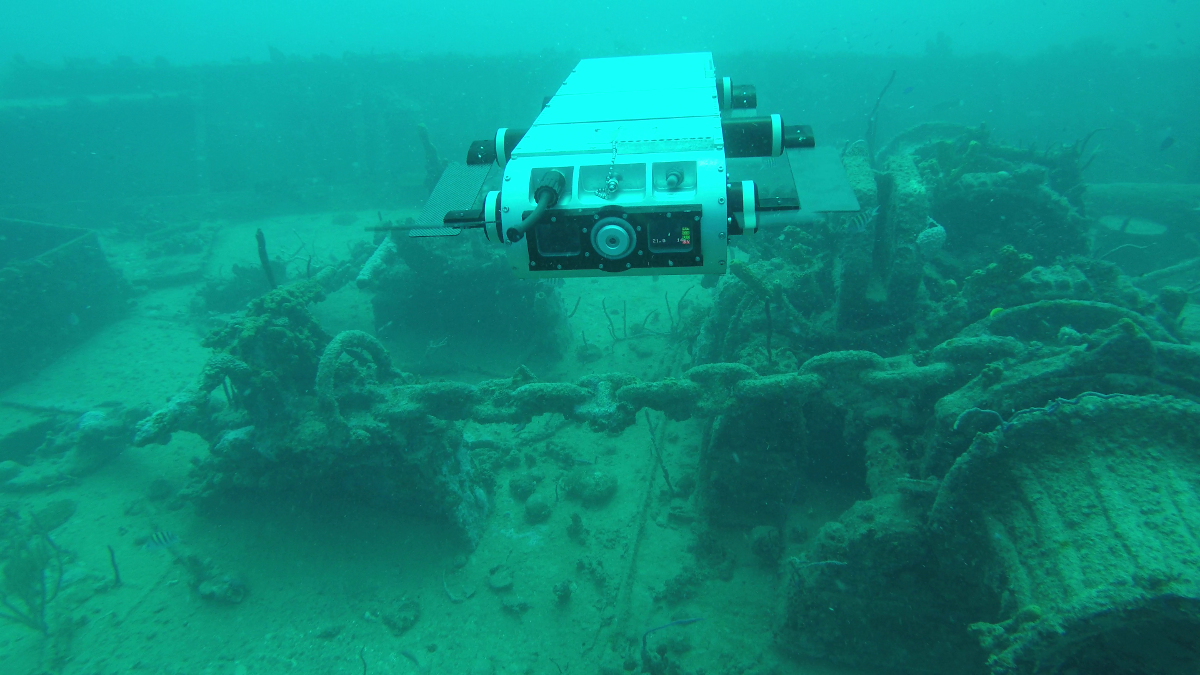}}&
        \subfigure[]{\includegraphics[height=0.13\textheight]{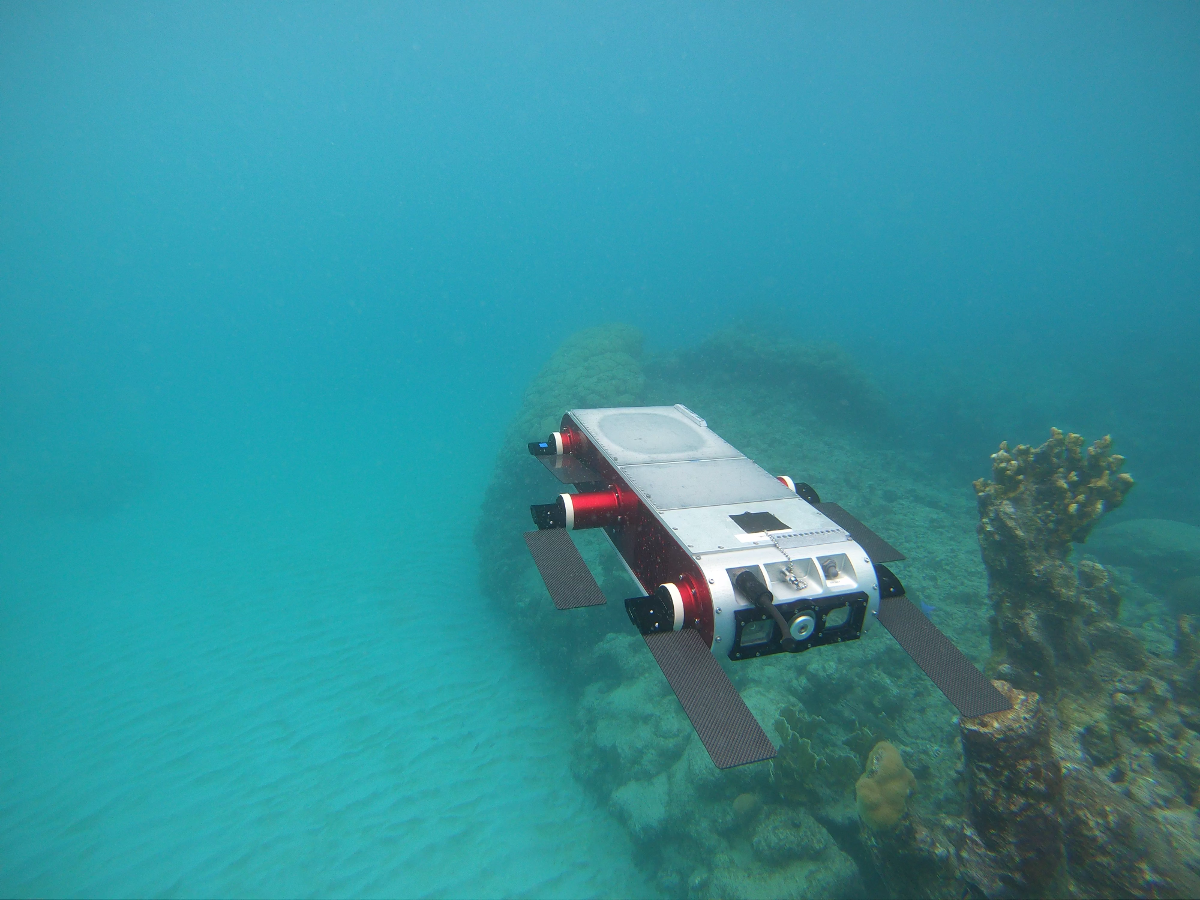}}&
        \subfigure[]{\includegraphics[height=0.13\textheight]{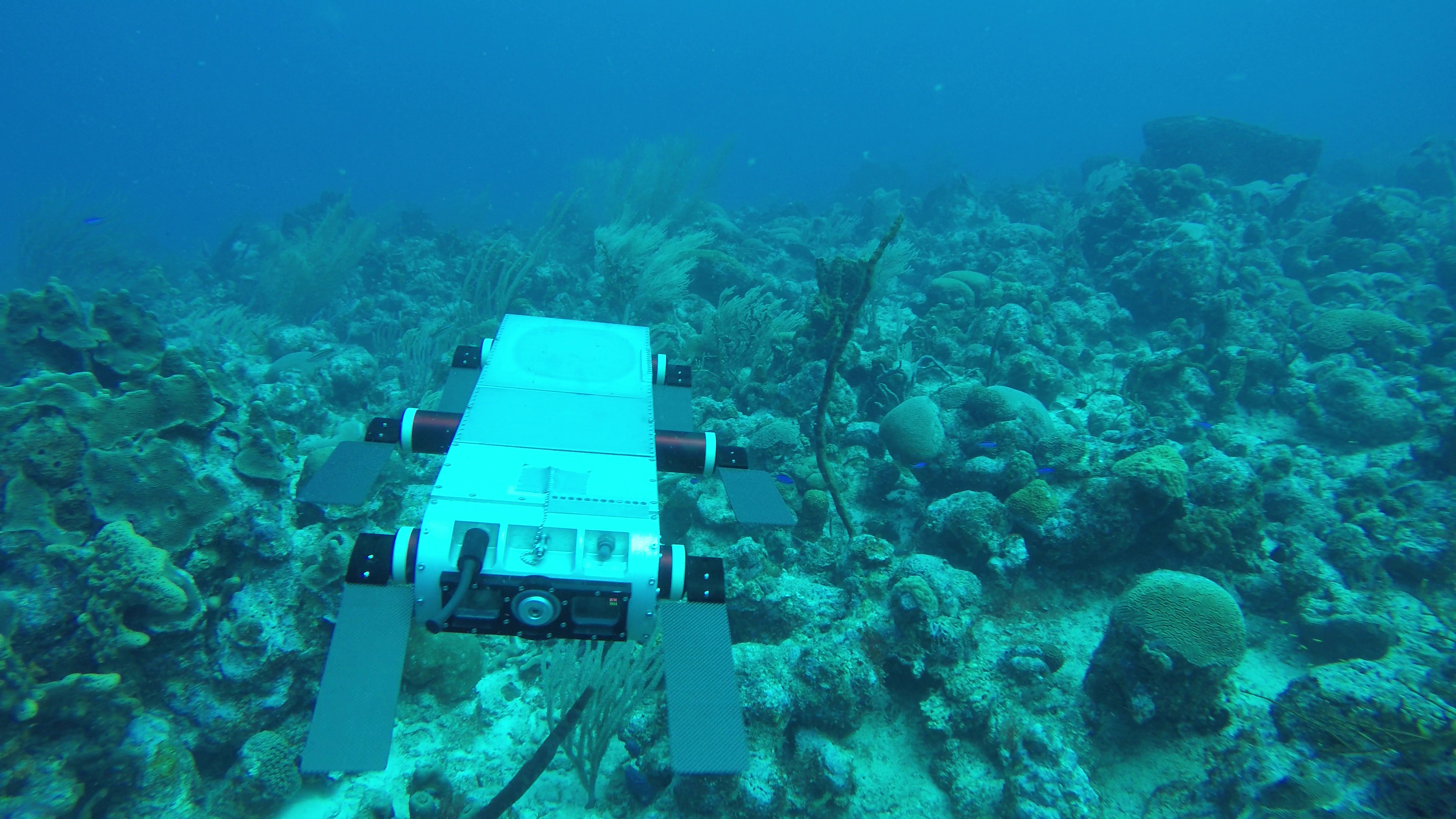}} 
    \end{tabular}
    \caption{Three environments where the  AUV was deployed (Barbados):  (a) over a  shipwreck performing a lawnmower pattern; (b) over a mixed sand and coral area performing multiple squares; (c) over a coral reef performing a lawnmower pattern.}
    \label{fig:environments}
\end{figure*}

When the VIO frontend can not detect or track enough keypoints to initialize a new keyframe, no keyframe information is generated. In this case, we wait for the specified time (set as a parameter) if we do not receive any keyframe information from SVIn2 for $kf\_wait\_time$ (generally set between 1 to 3 secs), we switch to the primitive estimator. Furthermore, we need to introduce these keyframes into the pose graph differently than regular keyframes as they only contain the odometry information from PE. These keyframes cannot be used for loop closure as they do not possess the keyframe image, features, and the 3D keypoints (used for geometric verification using PnP RANSAC) associated with them. It is worth noting that even if the SVIn2 health status is bad, we can use the keyframes originated from SVIn2 for loop closure.
\invis{
When this happens, $_{sv}\textbf{T}_{g}$ is reset to
\begin{equation}
    {}_{sv}\textbf{T}_{g} \coloneqq {}_{sv}\textbf{T}_{g} \cdot {}_W\textbf{T}_{sv,old} \cdot {}_W\textbf{T}_{pr,old}\inv \cdot {}_W\textbf{T}_{pe} \cdot {}_W\textbf{T}_{sv}\inv
\end{equation}
where ${}_W\textbf{T}_{sv,old}$ and ${}_W\textbf{T}_{pr,old}$ are the latest SVIn2 and PE poses, respectively, from a common timestamp before the failure.
}

\section{EXPERIMENTS}\label{sec:experiments}
\paragraph{\bf Datasets} \label{subsec:datasets}
The Aqua2 AUV has been deployed in a variety of challenging environments including shipwrecks, see \fig{fig:environments}(a); areas with sand and coral heads, see \fig{fig:environments}(b); and coral reefs, see \fig{fig:environments}(c). During each deployment, Aqua2 performs predefined trajectory patterns while using the odometry information from the PE. We have tested our approach on the following datasets:
\begin{itemize}
    \item \textbf{lawnmower over shipwreck}: The Aqua2 AUV performing a lawnmower pattern over the Stavronikita shipwreck, Barbados. During operations around shipwrecks, a common challenge is the lack of features when the wreck is out of the field of view; for example, while mapping the superstructure, the AUV can move over the side of the wreck (see \fig{fig:example}(e-f)), thus facing the open water with no reference. Since VIO is not able to track while facing open water, the AUV's pose cannot be estimated correctly without using the PE. We obtain the ground truth trajectory for the section with the shipwreck in view by using COLMAP \cite{colmap}, scale enforced using the rig constraints.
    
    \item \textbf{squares over coral reef}: The Aqua2 AUV performed square patterns over an area with sand and coral heads, Barbados; see \fig{fig:environments}(b). During operations over coral reefs, drop\hyp offs present similar conditions as wrecks, where the vehicle is facing blue water or a sandy patch. In addition, patches of sand present feature\hyp impoverished areas where VIO fails. \invis{However, the VIO was able to keep track when the Contrast Limited Adaptive Histogram Equalization (CLAHE) was applied to the images.}
    
    \item \textbf{lawnmower over coral reef}:  The Aqua2 performed a lawnmower pattern over a coral reef, Barbados; see \fig{fig:environments}(c). During operations, the VIO was able to track successfully the whole trajectory. This dataset was later artificially degraded to simulate loss of visual tracking. Utilizing the consistent track produced by VIO as ground truth, a quantitative study of the switching estimator is presented. It's worth mentioning that COLMAP was not able to register images during strips with fast rotation; hence not used for ground truth.
\end{itemize}
 
 \invis{
\begin{figure}[h]
 \centering
{\includegraphics[width=0.8\columnwidth]{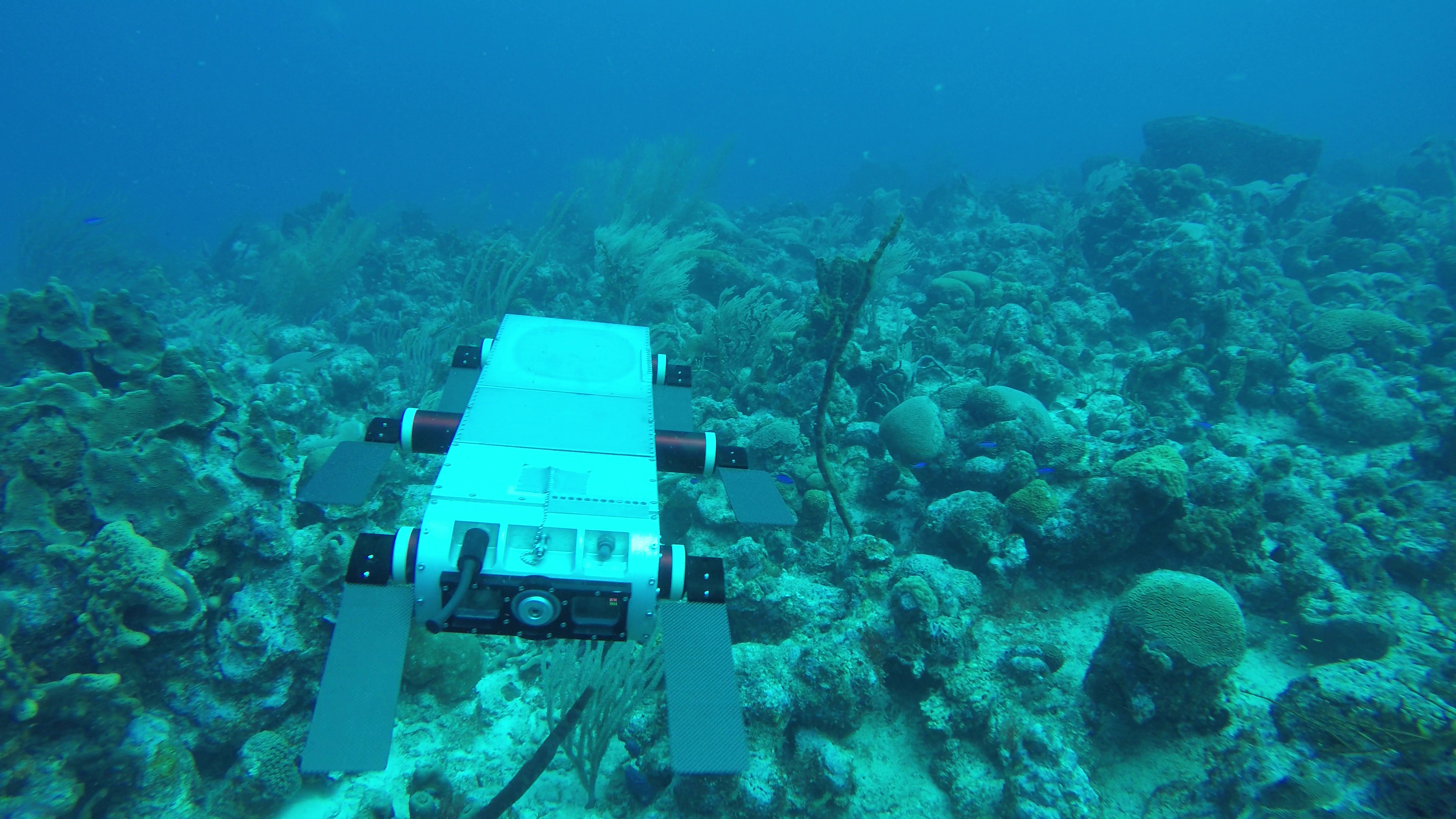}}
\caption{Aqua2 AUV navigating over the coral reef, Barbados.}
\label{fig:reef}
\end{figure}
}

\begin{figure*}[th]
    \centering
    \begin{tabular}{lcc}        
        \subfigure[]{\includegraphics[height=0.2\textheight]{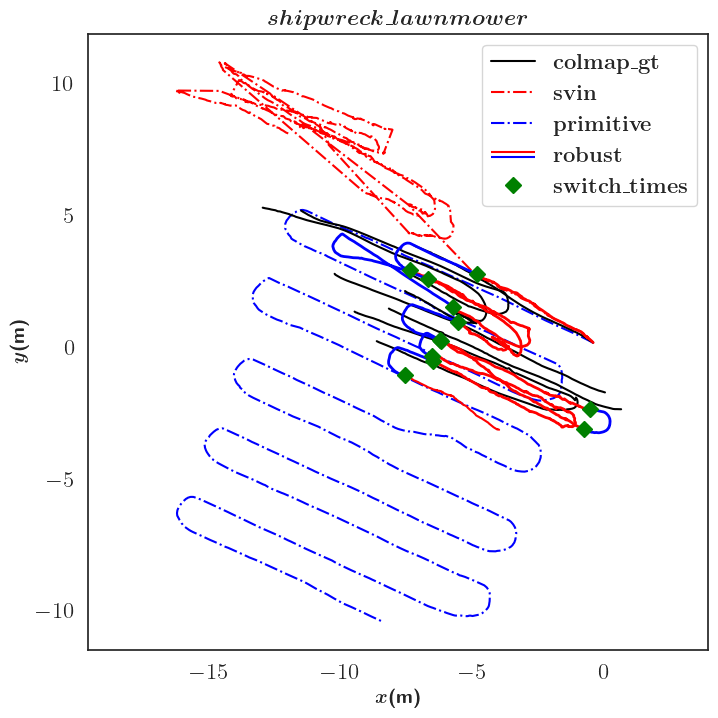}}&
        \subfigure[]{\includegraphics[height=0.2\textheight]{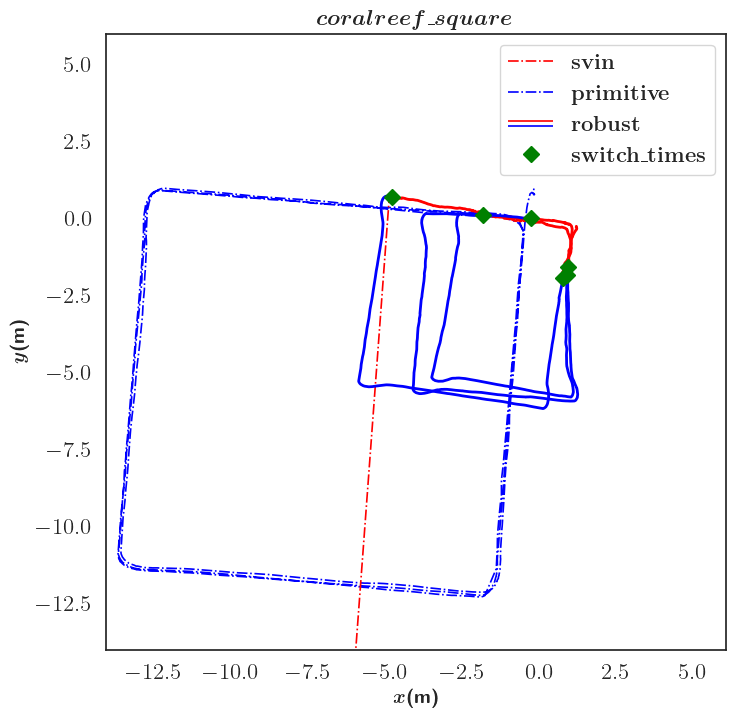}}&
        \subfigure[]{\includegraphics[height=0.2\textheight]{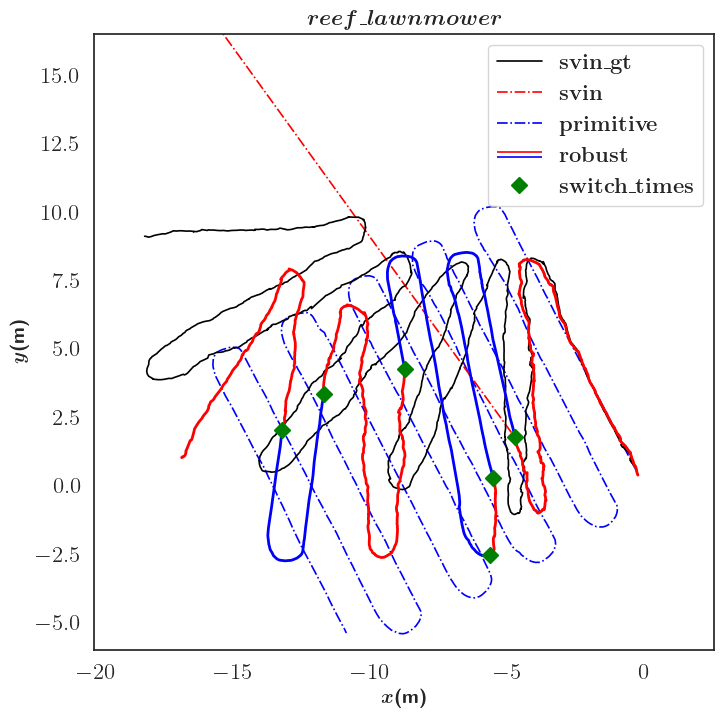}} 
    \end{tabular}
    \vspace{-0.1in}\caption{Resulting trajectories from three datasets. Each plot presents the PE trajectory, the SVIn2 trajectory, and the proposed robust switching estimator (solid line with red the parts of VIO and blue the PE contributions, switching points are marked by green points), approximate ground truth in (a) and (c) is plotted as a solid black line: (a) Stavronikita shipwreck, lawnmower pattern. (b) Mixed sand and coral area, multiple squares. Please note that as the SVIn2 trajectory lost track it moved far away. (c) Coral reef, lawnmower pattern. This dataset has no loop closures, however, SVIn2 maintained track over the complete trajectory. The visual data were artificially degraded  at random occasions to trigger the switch to PE.}
    \label{fig:trajectories}
\end{figure*}

\paragraph{\bf Trajectory Estimation} \label{subsec:results}
Trajectories were produced with  PE, SVIn2, and the proposed SM/VIO estimators. Figure \ref{fig:trajectories} presents the resulting trajectories for the three datasets. In all cases the PE trajectory (blue dash-dotted line) accurately traced the requested pattern as the primitive estimator was also used to guide the robot. The VIO (SVIn2) (red dash-dotted line) diverged when visual tracking failed. Finally the proposed estimator SM/VIO (solid red and blue line with green diamonds marking the switching of estimators) tracked consistently the pose of the AUV. \invis{Next, the results from the three datasets are discussed.}

The shipwreck\_lawnmower dataset presents a very challenging scenario, the AUV swims over the deck, VIO tracks consistently the feature\hyp rich clutter (\fig{fig:example}(d)), then the AUV approaches the sides of the wreck, the number of features is reduced (\fig{fig:example}(e)) until it goes over the side (\fig{fig:example}(f)) and faces blue water. As the detected features are drastically reduced the VIO continues forward, moving further away from the true position. It is worth noting that several loop closures kept the VIO estimate close enough to the wreck structure but in the wrong area. The proposed framework switched to the PE upon loss of visual tracking as can be seen from the green diamond signifying the switch in \fig{fig:trajectories}(a). COLMAP was able to register images in sections with shipwreck in view.


In the reef\_square dataset the AUV performed three squares over an area with some coral heads and a large sandy patch. As can be seen from \fig{fig:environments}(b) and \fig{fig:trajectories}(b) only one side of the square contains enough features for VIO tracking; however, these features enabled repeated loop closures. The primitive estimator over estimated the forward velocity producing squares much larger than the actual trajectory. The VIO upon loss of visual tracking failed (red dash-dotted line). SM/VIO produced accurate trajectories utilizing the loop closures. The top side of the square, where VIO was operational produced consistent trajectories across all  squares. The last dataset is discussed  next,  presenting a quantitative evaluation of the SM/VIO estimator.

\paragraph{\bf Quantitative Analysis}
The third dataset (lawnmower over a coral reef) produced VIO results without any loss of tracking, albeit without any loop closures. The visual input was artificially degraded (Gaussian blur with kernel size 21 and standard deviation 11 was introduced on selected images) randomly in order to generate controlled failures for the VIO. \fig{fig:trajectories}(c) presents such a scenario of three failures of 30 seconds each. For this study, these failures lasted for varying duration and a different number was introduced each time. More specifically, as can be seen in \tab{tab:RMSE}, we introduced one, three, and five failures, in a trajectory of 314 seconds with a total length of 108.13 meters as estimated by the successful SVIn2 estimator. Each scenario was run five times, the average Root Mean Square Error (RMSE) and standard deviation are reported. One failure of 60 seconds was introduced resulting in average RMSE of 3.2 meters. Three failures for 15, 30, and 45 seconds were introduced, resulting on average around 3 meters\invis{ per second}. Finally, five failures of 20 seconds were introduced resulting on average of RMSE 4.37 meters.   It is worth noting that in all cases the pure VIO  estimates diverged rapidly upon loss of visual tracking; see \fig{fig:trajectories}(c) red dash-dotted line. 

\begin{center}
\begin{table}[htb]
\centering
\invis{\caption{Lawnmower pattern over the coral reef (\fig{fig:trajectories}(c)), artificially degraded images to induce loss of tracking: RMSE between the SVIn2 trajectory and the Robust Switching Estimator. Mean and standard deviation over 5 trials.\label{tab:RMSE}}}
\caption{Quantitative analysis of robust switching estimator based on root mean squared translation error. The table shows mean and standard deviation of error over 5 runs.
\label{tab:RMSE}}
\begin{tabular}{@{}lccc}
  \toprule
   dataset & length & mean rmse & s.d. \\ 
   & (in meters) & (in meters) & (in meters)\\
\invis{  \midrule
  reef\_square & 53.15 & \num{2.059} & \num{0.258} \\}
  \midrule
  reef\_lmw\_1\_60 & 108.13 & 3.21 & 0.47 \\
  reef\_lmw\_3\_15 & 108.13 & 3.21 & 0.61 \\
  reef\_lmw\_3\_30 & 108.13 & \num{3.01} & \num{0.58} \\
  reef\_lmw\_3\_45 & 108.13 & \num{3.56} & \num{0.32} \\
  reef\_lmw\_5\_20 & 108.13 & 4.37 & 0.90\\
  \bottomrule
\end{tabular}
\end{table}
\end{center}

\invis{
\begin{figure}
 \begin{center}
 \includegraphics[width=0.98\columnwidth]{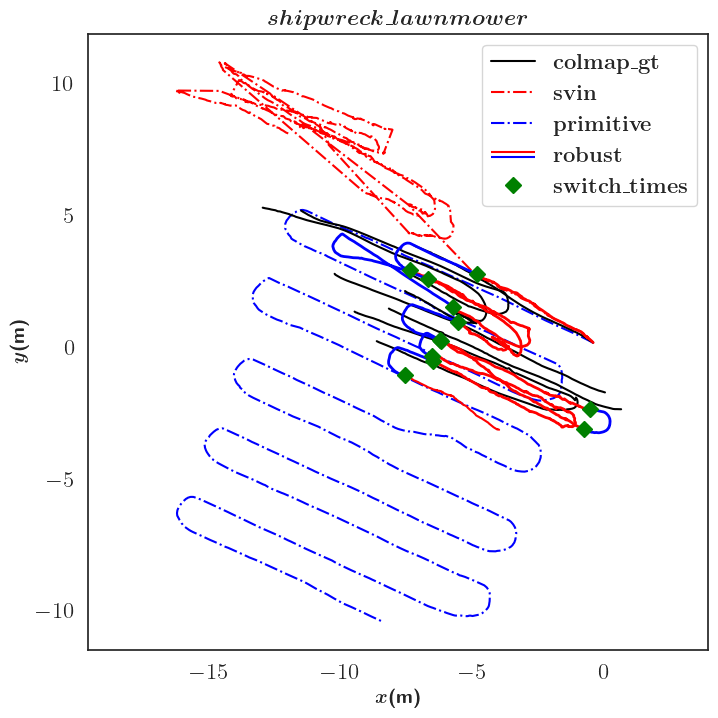}
\caption{The trajectories produced using SVIn2 and primitive estimator (up to scale) in the $shipwreck\_lawmmower$ dataset. The combined trajectory (dotted) is produced by combined  SVIn2 and primitive trajectories at the switching point using correct scale from SVIn2. Loop closure is disabled. }
\label{fig:switching_lmw}
 \end{center}
\end{figure}

\begin{figure}
 \begin{center}
 \includegraphics[width=0.98\columnwidth]{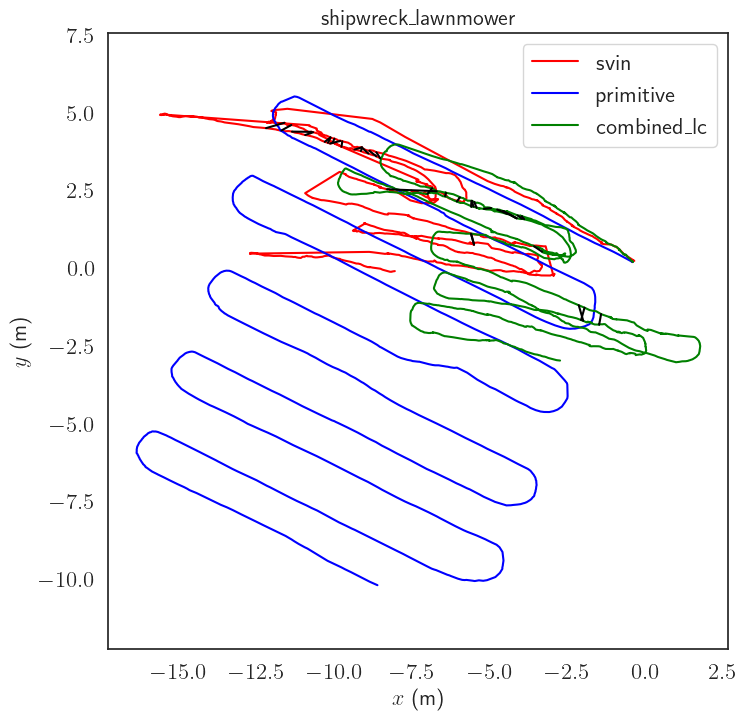}
\caption{The trajectories of SVIn2 with loop closure, primitive estimator and combined trajectory with loop closure. The perfectly straight trajectory section in SVIn2 are due to jump in trajectory. The black lines represent loop closure connections. \bcomment{may be plotting scaled primitive estimator would }}
\label{fig:result_shipwreck}
 \end{center}
\end{figure}
}
\vspace{-0.25in}
\paragraph{\bf Comparison with other VIO packages} The shipwreck\_lawnmower dataset was used to compare with well known VIO packages~\cite{Geneva2020ICRA, qin2019b, RahmanIROS2019a, leutenegger2015keyframe}. The ground truth is obtained using COLMAP~\cite{colmap} which was able to track images with shipwreck in view as it does not require continuous tracking. We compared the performance of various VIO algorithms with COLMAP baseline using root mean squared average translation error (ATE) metric after se3 alignment. As can be seen in \ref{tab:comp}, the proposed estimator maintained a pose estimate and exhibited the least RMSE, in contrast other algorithms deviated after losing track. OpenVINS \cite{Geneva2020ICRA} was not able to recover after losing track when the shipwreck is out of view and has a very high error. It is worth noting that all the VIO algorithms lose track when the robot approaches the side of the wreck facing blue water.

\begin{center}
\begin{table}[htb]
\caption{Performance of popular open\hyp source VIO packages on the wreck dataset. The root mean squared ATE compared to COLMAP trajectory after se3 alignment.}
\label{tab:comp}
\begin{tabular}{@{}lccc}
   \toprule
   VIO Algorithm & Time to first   &  Recovery? & RMSE \\
    &  track loss (in sec) &    &  (in m)\\
   \midrule
   OpenVINS \cite{Geneva2020ICRA} &  23.7 &  No  &  $\times$ \\
   OKVIS \cite{leutenegger2015keyframe} & 23.4 &   Partial & 5.199  \\
   VINS-Fusion\cite{qin2019b} & 23.6 &   Partial & 53.189 \\
   SVIn2\cite{RahmanIROS2019a} & 23.4   & Yes & 1.438 \\
   \bf SM/VIO & N/A &  Yes & 1.295  \\
   \bottomrule
\end{tabular}
\end{table}
\end{center}

\vspace{-0.35in}\section{CONCLUSION}
\label{sec:conclusions}
The presented estimator robustly tracked an AUV even when traveling through blue water or over a featureless sandy patch. The proposed system uses an Aqua2 vehicle~\cite{DudekIROS2005} and the SVIn2~\cite{RahmanIROS2019a} VIO approach; however, any AUV with a well-understood motion model can be utilized together with any accurate VIO package. \invis{In the future we plan to extend this approach to the BlueROV2 vehicle.} Recent deep learning based inertial odometry approaches ~\cite{neural_inertial_localization, rio, ronin} can also serve as a conservative alternative estimator. An evaluation of visual features for the underwater domain~\cite{QuattriniLiOceans2016,QuattriniLiICRAWork2016,ShkurtiCRV2011} will contribute additional information to the VIO health monitor.

\invis{Furthermore, evaluating measures of quality and quantity of the detected visual features for the underwater domain~\cite{QuattriniLiOceans2016,QuattriniLiICRAWork2016,ShkurtiCRV2011} in order to quantify the effectiveness of the visual component of the state estimation.}

Future use of the proposed approach will be to combine it with coral classification algorithms~\cite{ModasshirOceans2018,beijbom2015towards} in order to extract accurate coral counts over trajectories~\cite{ModasshirFSR2019} and models of the underlying reef geometry, and for mapping underwater structures~\cite{XanthidisISRR2022}. We are currently working on extending the Aqua2 vehicle operations inside underwater caves\invis{; see \fig{fig:cave}}. The challenging lighting conditions in conjunction with the extreme environment require the localization abilities of the vehicle to be robust even when one of the sensors fails temporarily. \invis{Finally, we will investigate the effect of VIO initialization methods for re-initialization after VIO health is bad. Finally, the developed framework of switching estimators in the demanding underwater domain will be transferred to other challenging scenarios, such as aerial and underground vehicles.}

\invis{
 \begin{figure}[h]
 \centering
{\includegraphics[width=0.8\columnwidth]{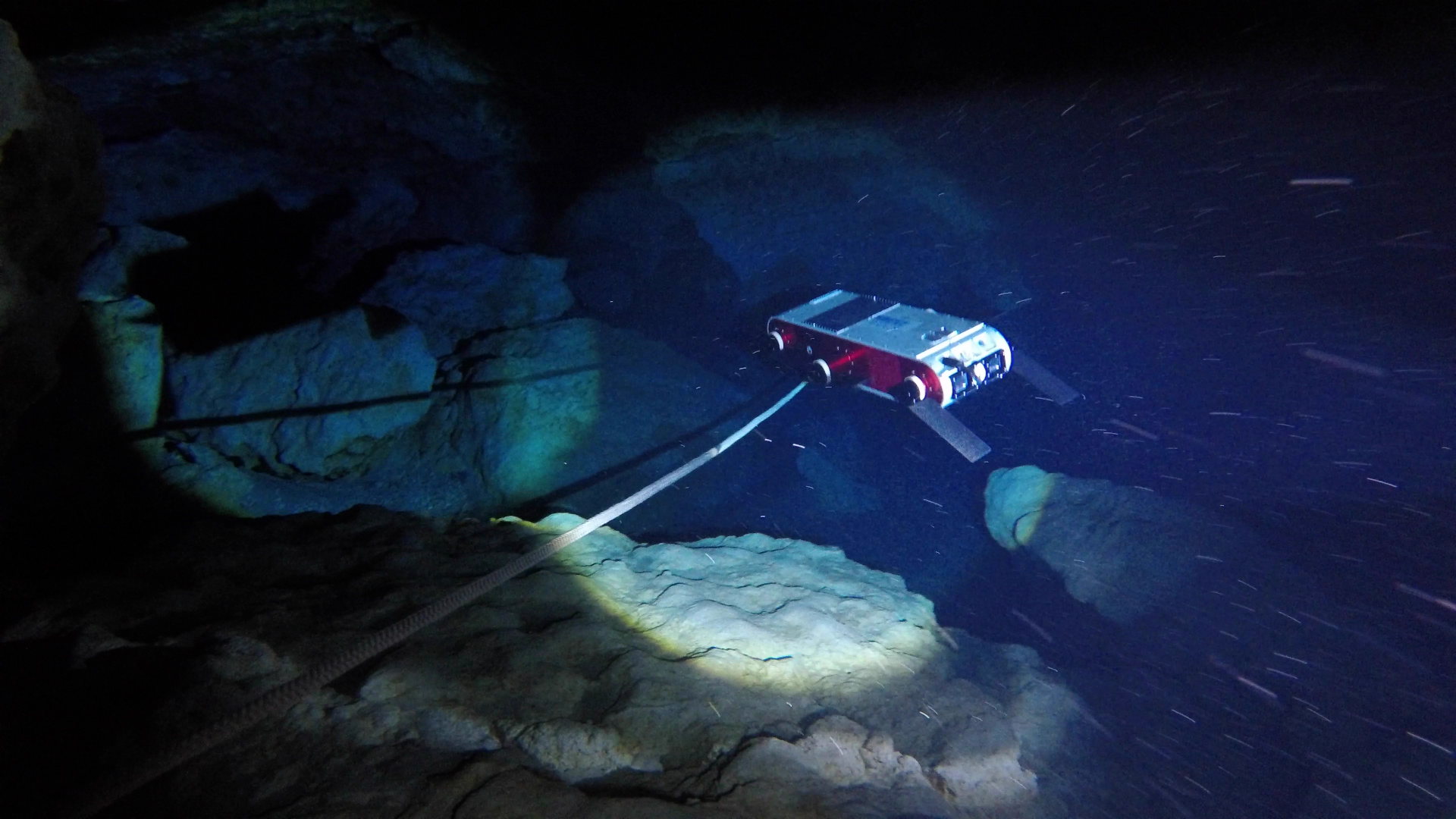}}
\caption{Aqua2 AUV operating inside a Cave, Ginnie Springs, FL.}
\label{fig:cave}
\end{figure}
}
\newpage
\newpage

%
%
\bibliographystyle{IEEEtran}
\bibliography{IEEEabrv,refs,pubs}

\end{document}